%% file: main.tex
\documentclass[11pt,a4paper]{article}
\usepackage[utf8]{inputenc}
\usepackage[T1]{fontenc}
\usepackage{amsmath,amssymb}
\IfFileExists{stix2.sty}{\usepackage{stix2}}{\usepackage{txfonts}}
\usepackage{microtype}
\usepackage[a4paper,margin=25mm,headheight=14pt]{geometry}
\usepackage{booktabs,array,calc}
\usepackage{graphicx}
\setkeys{Gin}{keepaspectratio}
\usepackage{float}
\makeatletter\renewcommand{\fps@figure}{!ht}\makeatother  
\usepackage{xcolor}
\definecolor{linknavy}{RGB}{20,50,110}
\usepackage[font=small,labelfont=bf,labelsep=period]{caption}
\usepackage{titlesec}
\titleformat*{\section}{\large\bfseries}
\titleformat*{\subsection}{\normalsize\bfseries}
\titleformat*{\subsubsection}{\normalsize\itshape}
\titlespacing*{\section}{0pt}{1.6ex plus .6ex}{0.8ex}
\titlespacing*{\subsection}{0pt}{1.2ex plus .4ex}{0.5ex}
\usepackage{enumitem}\setlist{nosep,leftmargin=*}
\usepackage{listings}
\usepackage{fancyhdr}
\usepackage{natbib}
\usepackage{lineno}
\newif\ifreview\reviewfalse   
\usepackage{authblk}

\usepackage{hyperref}
\usepackage[capitalise,noabbrev]{cleveref}  
\Crefname{figure}{Figure}{Figures}\Crefname{table}{Table}{Tables}\Crefname{section}{Section}{Sections}
\Crefname{appendix}{Appendix}{Appendices}\Crefname{subappendix}{Appendix}{Appendices}
\hypersetup{colorlinks=true,allcolors=linknavy,
  pdftitle={Derivative computation in physics-informed neural networks: automatic differentiation, finite differences and beyond},
  pdfauthor={Maciej J. Mikulski, Tadeusz Uhl}}
\newcommand{\passthrough}[1]{#1}

\title{\bfseries Derivative Computation in Physics-Informed Neural Networks:\\Automatic Differentiation, Finite Differences and Beyond}
\author[1,*]{Maciej J. Mikulski}
\author[2]{Tadeusz Uhl}
\affil[1]{AGH University of Krakow, Krakow, Poland; ORCID 0009-0001-6634-7934}
\affil[2]{AGH University of Krakow, Krakow, Poland; ORCID 0000-0002-4332-3067}
\affil[*]{Corresponding author: \texttt{mjmikulski@agh.edu.pl}}
\date{}

\begin{document}
\maketitle
\thispagestyle{fancy}
\ifreview\linenumbers\fi

\begin{abstract}

\noindent The physics in a physics-informed neural network (PINN) is only as reliable as its derivatives.
We investigate finite differences (FD) as an alternative to automatic differentiation (AD),
derive step-size scaling across derivative orders and floating-point formats,
and introduce periodic runtime calibration. Across four PDE benchmarks, calibrated FD matches AD accuracy;
for 4th-order plate bending in FP64, it trains 3× faster with 5.5× lower peak GPU memory.
We further identify a silent failure of the standard batched-autograd idiom when collocation points
interact through batch normalization or self-attention. For batch normalization the coupling can be removed
at the layer level; for self-attention the problem remains open. Together, these findings show that derivative computation
is both a performance opportunity and a correctness constraint in PINNs.


\medskip\noindent\textbf{Practitioner points}
\begin{itemize}
\item Finite differences reproduce the accuracy of automatic differentiation in PINNs on second- and fourth-order problems, and at fourth order at a fraction of the time and memory.
\item $\varepsilon$ calibration is crucial and should be repeated a few times during training.
\item The \texttt{grad\_outputs=ones} idiom gives wrong derivatives for networks with batch normalization or self-attention over collocation points; check per-sample derivatives before trusting the residual.
\end{itemize}
\end{abstract}

\begin{figure}[!ht]
\centering
\includegraphics[width=\linewidth]{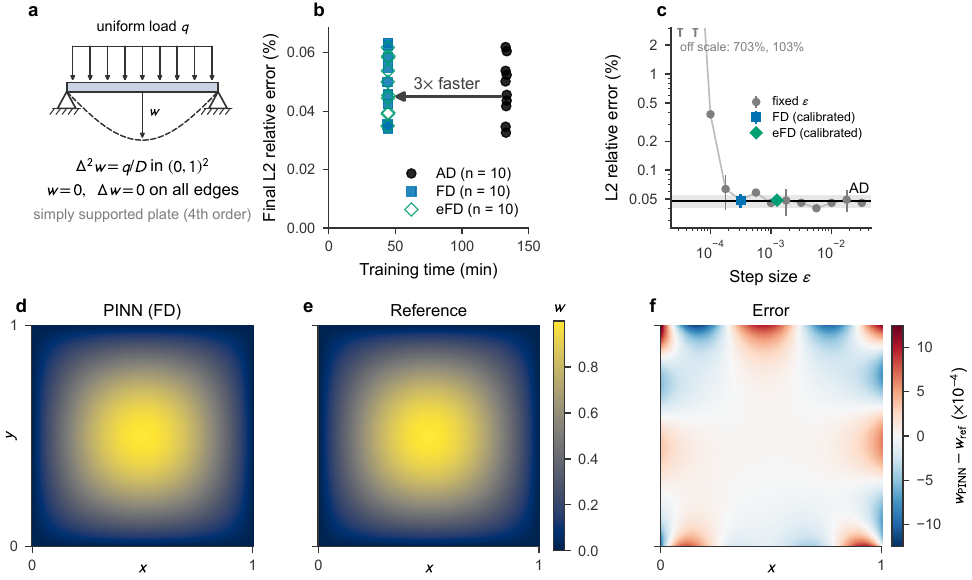}
\caption{Bending of a simply supported plate under a uniform load, solved by a Physics-Informed Neural Network whose
PDE residual is computed with finite differences (FD; \Cref{sec:plate}).
(a) The problem definition.
(b) Final L2 relative error against training time, one point per run: FD and eFD reach the accuracy of nested automatic differentiation (AD) about three times faster.
(c) L2 relative error of FD with a fixed step $\varepsilon$ held for the whole run (grey; whiskers: 95\% CI over five seeds where replicated)
against AD (line and band: mean $\pm$ 95\% CI), with the calibrated steps of FD and eFD.
(d) PINN prediction of the deflection $w$ (FD with the calibrated step, one training run), (e) reference (Navier series),
(f) pointwise error.}
\label{fig:teaser}
\end{figure}

\input{s1_introduction}
\input{s2_related_work}
\input{s3_method}
\input{s4_experiments}

\input{s5_discussion}
\input{s6_declarations}
\appendix
\input{s7_appendix}

\bibliographystyle{plainnat}
\bibliography{refs}
\end{document}

%% file: s1_introduction.tex
\hypertarget{introduction}{%
\section{Introduction}\label{introduction}}

Partial differential equations (PDEs) underpin the quantitative
description of heat transfer, wave propagation, structural deformation
and fluid flow. Mesh-based solvers such as the finite-element and
finite-volume methods are mature and robust, but mesh generation is
laborious for complex geometries, and inverse problems require
substantial reformulation. Physics-Informed Neural Networks (PINNs)
\citep{raissi2019pinn, Cuomo2022Scientific} address both by embedding
the governing equations in the loss function of a neural network, which
handles sparse or noisy data, inverse problems and irregular geometries
without a mesh \citep{DeRyck2024Numerical}. The price is that they
compete with decades of progress in numerical algorithms
\citep{Brunton2024Promising}, so the numerical quality of their building
blocks, derivative computation included, is a first-class concern.

PINNs face several practical challenges: balancing loss terms of
different scales, choosing collocation points, activation functions
\citep{Jagtap2020Adaptive} and architectures. Among these, the
computation of derivatives is particularly critical: the PDE residual
requires partial derivatives of the network output with respect to its
inputs, often of second or higher order. Fourth-order operators are
common in computational mechanics (Kirchhoff--Love plates and shells,
strain-gradient elasticity, phase-field and Cahn--Hilliard models), and
the cost of nested automatic differentiation grows quickly with the
derivative order.

Modern deep learning frameworks such as PyTorch
\citep{paszke2019pytorch} and JAX \citep{jax2018github} provide
automatic differentiation (AD) that, in principle, makes derivative
computation straightforward. These mechanisms are, however, designed and
optimized for the gradient of a scalar loss with respect to the network
weights, as needed in backpropagation. Derivatives of the output with
respect to the inputs at every collocation point are obtained through a
batched trick (a vector--Jacobian product with a vector of ones) that is
exact only for architectures without coupling between samples. In
practice, AD-based residuals also lead to a rapid growth of the
computational graph, especially for second and higher derivatives such
as the Laplacian, which becomes prohibitive for large batch sizes or
high-order equations.

Furthermore, certain architectural choices exacerbate these
difficulties. Batch normalization (BN), widely used to stabilize
training, and self-attention, the building block of Transformers
\citep{vaswani2017attention}, introduce dependencies between samples
within a mini-batch or a sequence. When computing per-sample
derivatives, as required for the PDE residual at each collocation point,
these inter-sample dependencies lead to incorrect derivatives if not
handled with care. To the best of our knowledge, we are the first to
signal this problem.

Given these limitations, there is renewed interest in finite-difference
(FD) methods for derivative approximation within neural network training
\citep{lim2022fdm}. While FD is among the oldest techniques in numerical
analysis, its application to modern deep learning has been largely
overlooked in favor of AD. Recent work has demonstrated its viability in
large-scale settings: \citet{wang2025tim} replaced backpropagation-based
Jacobian computation with FD in a large generative model, achieving over
2\(\times\) speedup. FD requires only forward function evaluations,
which parallelize trivially, rather than the sequential graph traversals
inherent to AD. This makes it particularly attractive when higher-order
derivatives are needed.

In this work, we present a systematic benchmark study of finite
differences for derivative computation in Physics-Informed Neural
Networks. Our contributions are as follows.

First, we demonstrate that the step size \(\varepsilon\) is crucial for
numerical accuracy and we propose a simple empirical heuristic for
selecting \(\varepsilon\) based on the interplay between truncation and
floating-point errors. We introduce a scheme that periodically
recalibrates \(\varepsilon\) during training as the network's smoothness
characteristics evolve.

Second, we provide an analysis of how the optimal \(\varepsilon\) scales
with the derivative order and propose two practical strategies for
mixed-order PDEs: a geometric-mean step size (FD) and order-specific
step sizes (eFD). We verify them on three 2nd-order benchmark PDEs and
on a 4th-order plate-bending problem (\Cref{fig:teaser}). We also
propose and explore a new stochastic step-size variant (sFD).

Additionally, we note that a common way to compute derivatives for
neural networks containing inter-sample dependencies (e.g.~BN or
self-attention) leads to silently incorrect results, and provide
theoretical and experimental evidence.

Our experiments on four benchmark PDEs, up to fourth order (steady and
transient heat conduction, shock formation in a fluid and thin-plate
bending), demonstrate that FD-based PINNs achieve the accuracy of their
AD-based counterparts while reducing memory consumption and enabling
larger batch sizes; at fourth order the cost advantage is largest, see
\Cref{fig:teaser}. While \citet{Chen2024Automatic} recently argued that
AD is essential for training neural networks to solve differential
equations, our results show that FD is interchangeable with AD for
standard MLP-based PINNs in practical settings.

These findings suggest that finite differences, far from being an
obsolete technique, offer a viable and often preferable alternative for
derivative computation in physics-informed deep learning.

%% file: s2_related_work.tex
\hypertarget{related-work}{%
\section{Related Work}\label{related-work}}

\hypertarget{finite-differences-in-pinns}{%
\subsection{Finite Differences in
PINNs}\label{finite-differences-in-pinns}}

The idea of replacing automatic differentiation with finite-difference
formulas in PINNs has been explored by several groups.
\citet{lim2022fdm} proposed a straightforward substitution of AD with
central differences for computing PDE residuals, demonstrating
feasibility on simple benchmark problems. \citet{jiang2023fdpinn}
applied FD-PINNs to steady incompressible flow problems and reported
higher accuracy, faster convergence, and greater robustness compared to
AD-based PINNs; in their lid-driven cavity benchmark at high Reynolds
numbers, the FD-PINN outperformed both the AD-PINN and a classical CFD
solver on a coarse mesh. \citet{roy2025fdns} confirmed these findings on
Navier--Stokes problems, showing that FD-PINNs can match or exceed
AD-PINN accuracy while reducing computational cost.
\citet{zhang2024rk4pinn} further demonstrated that
finite-difference-based physics-informed deep learning can achieve
efficient and accurate PDE solutions. \citet{Ha2022Physics-Informed}
combined numerical differentiation with physics-informed training for
complex fluid dynamics, and \citet{Guo2025FDPINN} proposed an FD-PINN
with adaptive residual point selection for seismic inversion, coupling
FD derivatives with spatially adaptive sampling.

On the theoretical side, \citet{langer2026illposed} proved that both
AD-PINNs and FD-PINNs are mathematically ill-posed (minimizers are
non-unique), but that FD-PINNs are ``tightly coupled'' to the underlying
FD scheme: any zero-loss FD-PINN solution on a fixed grid corresponds to
a true solution of the discrete PDE on that grid. That motivates looking
at FD-PINNs as full citizens of the numerical analysis world, rather
than as an approximation to AD-PINNs.

\citet{DeRyck2024Numerical} provided a comprehensive numerical analysis
of PINNs and related models, placing both AD and FD approaches within a
rigorous approximation-theoretic framework, addressing the problem of
the existence of the FD operator at the edges of the domain.

\hypertarget{hybrid-approaches}{%
\subsection{Hybrid Approaches}\label{hybrid-approaches}}

Rather than choosing between AD and FD, several works combine both.
\citet{chiu2022canpinn} proposed CAN-PINN, where finite-difference-type
stencils couple neighboring support points while AD supplies local
derivative information inside the stencil. This was motivated by the
observation that standard AD-PINNs constrain residuals almost pointwise
at collocation points, which can be under-constraining in
sparse-sampling regimes.

\citet{xiang2022hfd} developed HFD-PINN, which uses AD near complex
boundaries (where FD stencils are difficult to construct) and FD in the
interior. Their variant HFD-PINN-sdf uses a signed distance function to
ensure stencil points remain inside the domain, enabling mesh-free
finite differences on irregular geometries.

\hypertarget{step-size-selection}{%
\subsection{Step Size Selection}\label{step-size-selection}}

Despite the importance of \(\varepsilon\), most existing FD-PINN works
treat it as a fixed hyperparameter chosen by trial and error.
\citet{wang2025tim}, in the context of generative diffusion models, used
a manually chosen small \(\varepsilon\) to approximate Jacobian-vector
products via forward differences, treating it as a constant throughout
training. \citet{lim2022fdm} use a finite-difference step size
determined by the grid resolution. In their Burgers equation example,
this notation obscures the fact that the spatial and temporal grids have
different spacings, since both are denoted by a single \(h\).
\citet{jiang2023fdpinn} and \citet{xiang2022hfd} likewise used fixed
step sizes.

To our knowledge, no prior work proposes (i) an empirical
runtime-calibration procedure for \(\varepsilon\) that adapts to the
current network state, (ii) periodic recalibration of \(\varepsilon\)
during training to track the evolving smoothness of the solution, or
(iii) order-specific or stochastic step-size sampling as a strategy for
mixed-order PDEs.

\hypertarget{fd-beyond-pinns-and-alternative-derivative-methods}{%
\subsection{FD Beyond PINNs and Alternative Derivative
Methods}\label{fd-beyond-pinns-and-alternative-derivative-methods}}

The computational benefits of finite-difference-style derivative
approximations are not limited to PINNs. \citet{Pang2020Efficient}
reformulated score matching in generative modeling by expressing
Hessian-trace-related terms as directional derivatives and approximating
them with finite-difference decompositions. This replaced sequential
higher-order AD computations with parallelizable function evaluations,
yielding a 1.7--2.9\(\times\) speedup and a 10--35\% reduction in memory
usage across various models and datasets.

In a billion-parameter generative model, \citet{wang2025tim} replaced
AD-based Jacobian-vector products (JVPs) with a forward-pass-only
finite-difference estimator, which is reported to be approximately
2\(\times\) faster than the JVP while being natively compatible with
fully sharded data parallel training. This result illustrates that
avoiding explicit AD-based derivative computation can improve not only
runtime, but also the scalability of training large neural models.

Beyond finite differences, several alternative derivative computation
strategies have been proposed for PINNs. \citet{Cardona2023Polynomial}
introduced polynomial differentiation (``Sobolev cubatures''), which
replaces nested AD calls with precomputed cubature matrices, yielding
substantial speedups. \citet{Sharma2022Accelerated} accelerated PINN
training using meshless RBF-FD discretizations that provide high-order
accurate derivatives on scattered point clouds. They also successfully
used a mixed strategy: RBF-FD for the spatial derivatives and AD for the
time derivative. A similar RBF approach was used successfully by
\citet{Xiao2023Radial}.

\citet{Cho2023Separable} proposed separable PINNs that decompose the
problem structure to reduce derivative computation costs.

\citet{Chen2024Automatic} argued that replacing AD with standard FD can
adversely affect the training dynamics of PINN-style losses. Their
spectral analysis links this effect to the distribution of small
singular/eigenvalues and introduces a truncated-entropy measure as a
proxy for late-stage convergence, suggesting that AD can achieve lower
residual errors and faster optimization. This result provides a useful
caution against treating FD as a universally benign drop-in replacement
for AD, although the practical relevance of this effect depends on the
discretization, architecture, derivative order, and whether performance
is measured by residual minimization, solution accuracy, memory usage,
or wall-clock time.

%% file: s3_method.tex
\hypertarget{sec:method}{%
\section{Methodology}\label{sec:method}}

\hypertarget{finite-differences-for-neural-networks}{%
\subsection{Finite differences for Neural
Networks}\label{finite-differences-for-neural-networks}}

For a given function, one can obtain an approximation of its derivative
at point \(x\) by evaluating the function at a few points in the
vicinity of \(x\). While there exist multiple schemes for such an
approximation, we focus here on 3-point central differences. For the
first derivative the scheme reads:

\[
D_\varepsilon^{(1)} f(x) = \frac{f(x + \varepsilon) - f(x - \varepsilon)}{2\varepsilon},
\] where \(D_\varepsilon^{(k)}\) denotes the finite-difference operator
of order \(k\) with step \(\varepsilon\). Similarly, the second
derivative can be obtained as:

\[
D_\varepsilon^{(2)} f(x) = \frac{f(x + \varepsilon) - 2f(x) + f(x - \varepsilon)}{\varepsilon^2}.
\]

Given a (physics-informed) neural network, i.e.~a mapping
\(\mathcal{D} \ni x \mapsto f_\theta(x) \in \mathbb{R}\), we use the
above schemes to approximate spatial derivatives of \(f_\theta(x)\) at
collocation points \(X = \{x_i\}_{i=1}^B \subset \mathcal{D}\). In
practice, this amounts to evaluating the neural network on an enlarged
batch of perturbed points, i.e.~the concatenation of
\([X-\varepsilon, X, X+\varepsilon]\). This contrasts with
\citet{lim2022fdm}, where a grid is fixed and \(\varepsilon\) is
determined by the grid spacing. Algorithm 1 provides pseudocode for the
one-dimensional case \(\mathcal{D}=\mathbb{R}\); the formulation for
\(\mathcal{D}=\mathbb{R}^d\) and the stencils of order 3 and 4 used in
\Cref{sec:plate} are given in \Cref{app:stencils}. A practical PyTorch
implementation will be made publicly available upon publication of this
paper.

Throughout, AD denotes the computation of the input derivatives that
enter the PDE residual; the gradient of the loss with respect to the
network weights is always obtained by backpropagation, for FD as well as
for AD, and both use the autograd engine of PyTorch.

\textbf{Algorithm 1: Finite-difference derivatives of a network output
(\passthrough{\lstinline!finite\_difference!})}

\begin{lstlisting}
METHOD finite_difference
    # Inputs:
    # model: neural network mapping R to R
    # X: batch of randomly sampled collocation points, shape (B,)
    # epsilon: finite-difference step size
    stencil_1 = [[-0.5, 0.0, 0.5]]  # 1st deriv coeffs
    stencil_2 = [[1.0, -2.0, 1.0]]  # 2nd deriv coeffs
    # Step 1: Create perturbed points
    X_plus = X + epsilon 
    X_minus = X - epsilon 
    # Step 2: Concatenate original and perturbed points
    X_concat = concatenate([X_minus, X, X_plus])
    # Step 3: Evaluate the network at all points in a single forward pass
    u_concat = model(X_concat)  
    # Step 4: Reshape to separate original and perturbed evaluations
    u = u_concat.view(3, B)
    # Step 5: Compute finite-difference approximations
    u_x = u @ stencil_1.T / epsilon            # 1st derivative
    u_xx =  u @ stencil_2.T  / (epsilon ** 2)  # 2nd derivative
    return u_x, u_xx
\end{lstlisting}

\hypertarget{sec:error-model}{%
\subsection{Error model and the optimal step
size}\label{sec:error-model}}

\begin{figure}
\hypertarget{fig:format-eps}{%
\centering
\includegraphics[width=1\textwidth,height=\textheight]{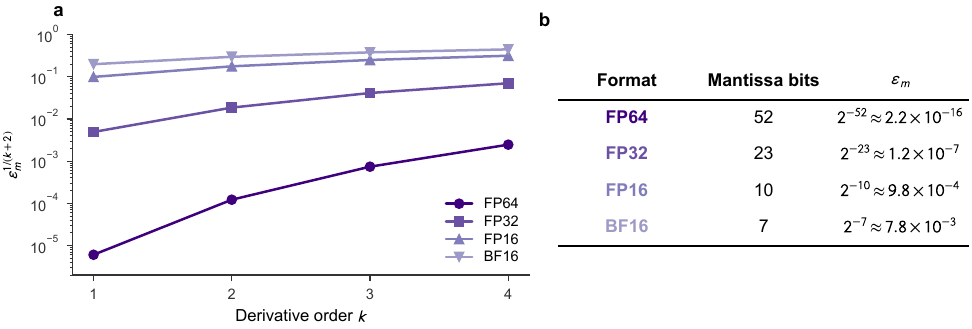}
\caption{Theoretical scaling of the optimal step with the derivative
order and the floating-point format: (a) \(\varepsilon_m^{1/(k+2)}\) for
\(k = 1\)--\(4\); the optimal step is \(C_k\) times this value; (b)
machine epsilon \(\varepsilon_m\) of the formats
shown.}\label{fig:format-eps}
}
\end{figure}

Finite differences involve a trade-off between the truncation error and
round-off error. The \textbf{truncation error} has a mathematical
origin: \(D_\varepsilon^{(k)}\) approximates the actual derivative using
only the first few terms of the Taylor expansion. Higher-order terms are
truncated, i.e.~neglected. The \textbf{round-off error} is a
computational error --- it arises from the limited number of bits we use
to represent a number in computer memory. Each floating-point format has
a given \textbf{machine epsilon} \(\varepsilon_m\), the spacing between
\(1\) and the next representable number. Differences of nearly equal
values, such as the numerators of finite-difference formulas, lose most
of their significant digits to rounding (cancellation), and the
subsequent division by a power of \(\varepsilon\) amplifies the error
that remains. Concrete values of the machine epsilon are listed in
\Cref{fig:format-eps}b.

For a derivative order \(k\) with 2nd-order accuracy, the two
contributions scale as (see derivation in \Cref{app:error})

\[
E^{(k)}(\varepsilon) \approx a_k\, \varepsilon^2 + b_k\, \frac{\varepsilon_m}{\varepsilon^k}, 
\] where \(a_k \propto |f^{(k+2)}(x)|\) and \(b_k \propto |f(x)|\). That
leads to a characteristic V-shaped curve of the total error as a
function of \(\varepsilon\), see \Cref{fig:vcurves}.

Minimizing over \(\varepsilon\) gives an optimal step size where the sum
of the two errors is minimal. It turns out that the optimal step size
depends on the machine epsilon and derivative order:

\[
\varepsilon_\mathrm{opt}^{(k)} = C_k \cdot \varepsilon_m^{1/(k+2)},
\] where \(C_k = \left( \frac{k\, b_k}{2\, a_k} \right)^{1/(k+2)}\).
Obviously, the location of the minimum is not fixed: \(C_k\) depends on
the ratio \(|f(x)| / |f^{(k+2)}(x)|\), which, when \(f\) is a neural
network being trained, evolves as the model's smoothness and
nonlinearity change during optimization. So \(C_k\) is in general
unknown. Nevertheless it is illustrative to see how orders of magnitude
change with the derivative order and how they differ between floating
point formats, see \Cref{fig:format-eps}a.

\Cref{fig:format-eps}a shows the scaling \(\varepsilon_m^{1/(k+2)}\) for
the derivative orders used in this paper and the common floating-point
formats; the actual optimum differs by the function-dependent prefactor
\(C_k\), which the analysis leaves undetermined --- a gap we close in
the next section with a simple empirical procedure for selecting
\(\varepsilon\) at runtime.

\hypertarget{sec:calibration}{%
\subsection{Empirical Determination of Optimal Step
Size}\label{sec:calibration}}

The above analysis motivates the following procedure, which evaluates
\(n\) logarithmically spaced candidate step sizes and selects the one
that minimizes the RMSE between the FD approximation and an AD-computed
reference.

\textbf{Algorithm 2: Empirical step-size calibration
(\passthrough{\lstinline!find\_eps!})}

\begin{lstlisting}
METHOD find_eps:
    # Inputs:
    #   model: neural network approximating function f
    #   order: derivative order k
    #   n_candidates: number of candidate step sizes, e.g. 50
    # Heuristic range for FP32:
    #   eps_min = 1e-6, eps_max = 1e-1
    eps_candidates = torch.logspace(-6, -1, n_candidates)
    X = Set of collocation points, e.g. from Sobol sequence
    # Compute AD reference (per-sample, exact)
    ref = ad_derivative(model, X, order)
    # Evaluate FD error at each candidate
    errors = []
    for eps in eps_candidates:
        fd_approx = fd_derivative(model, X, order, eps)
        errors.append(rmse(fd_approx, ref))
    return eps that minimizes error
\end{lstlisting}

We perform this calibration of \(\varepsilon\) at the beginning and then
a few times during training. \Cref{sec:interval} reports the measured
sensitivity to the recalibration interval.

\hypertarget{sec:mixed}{%
\subsection{Strategies for Mixed-Order PDEs}\label{sec:mixed}}

PDEs usually involve derivatives of multiple orders; for example the
viscous Burgers equation contains first- and 2nd-order terms, and the
plate problem requires a 4th-order operator with a 2nd-order boundary
condition. We therefore run Algorithm 2 separately for each derivative
order present, obtaining e.g.~\(\varepsilon_\mathrm{opt}^{(1)}\) and
\(\varepsilon_\mathrm{opt}^{(2)}\). The question arises: how to use
these different optima when performing finite-difference-based training?
We propose three FD-based strategies:

\textbf{Strategy 1: basic Finite Difference (FD)} Use a single
compromise step size computed as the geometric mean of the
order-specific optima, e.g.~for Burgers: \[
\varepsilon_{\mathrm{GM}} = \sqrt{\varepsilon_\mathrm{opt}^{(1)} \cdot \varepsilon_\mathrm{opt}^{(2)}}.
\] This requires only one set of function evaluations per stencil but
achieves suboptimal accuracy for both derivative orders.

\textbf{Strategy 2: Exact order-specific step sizes Finite Difference
(eFD)} Use \(\varepsilon_\mathrm{opt}^{(1)}\) for first derivatives and
\(\varepsilon_\mathrm{opt}^{(2)}\) for second derivatives, etc. This
achieves optimal accuracy for each term at the cost of a larger stencil
when two orders share a coordinate within the same stencil (on Burgers,
7 instead of 5 points per collocation point, i.e.~a 40\% larger
effective batch size); when the two orders enter separate loss terms, as
on Heat2D-CG (second order in the residual, first order in the Robin
condition), eFD costs the same as FD, and when each PDE term carries the
same derivative order, as e.g.~in the Poisson equation, eFD is
equivalent to FD.

\textbf{Strategy 3: Stochastic step sizes Finite Difference (sFD)} At
each collocation point use a different step size chosen randomly from
the suitable range.

We systematically compare these strategies in our experiments,
evaluating their impact on approximation accuracy, memory usage, and
wall-clock time across the mixed-order PDE benchmarks. A stochastic
variant is explored separately at the end of \Cref{sec:experiments}.

\hypertarget{sec:persample}{%
\subsection{Per-sample Derivatives in Architectures with Inter-Sample
Dependencies}\label{sec:persample}}

Most of the PINN literature quietly assumes that the neural network
\(f_\theta\) maps each input \(x_i\) independently to an output
\(f_\theta(x_i)\). Standard MLPs, when processing a batch of points in
parallel, satisfy this assumption. However, architectures that contain
layers with inter-sample dependencies break this assumption.

One class of such architectures consists of neural networks using the
default implementation of batch normalization \citep{ioffe2015bn}, a
common technique that is known to improve convergence of the
optimization. Another, even more interesting class, are architectures
with self-attention layers where the collocation points are not treated
independently, but influence each other in a highly non-linear manner.
Spectacular successes of transformer architectures in multiple domains
(e.g.~GPT \citep{radford2018gpt}) motivate research on models of this
kind, see e.g.~\citet{zhao2023pinnsformer}.

Unfortunately, these inter-sample couplings have consequences for
derivative computation. Reverse-mode AD computes vector--Jacobian
products, so element-wise derivatives are not directly available. The
common trick of setting the vector to ones (in PyTorch
\passthrough{\lstinline!grad\_outputs=ones\_like(outputs)!}) is silently
wrong: it returns the column sums \(\sum_i \partial u_i/\partial x_j\)
of the batch Jacobian instead of its diagonal
\(\partial u_j/\partial x_j\). This is not specific to one framework: a
forward-mode product with an all-ones tangent returns the row sums
\(\sum_j \partial u_i/\partial x_j\), and neither sum is the diagonal.
The correct approach is to compute the per-sample gradient with \(O(B)\)
separate autograd calls, one forward-backward pass per sample, which is
prohibitively expensive in time and memory at the batch sizes needed for
PINN training.

For BN, the problem may be simply overcome both for AD and FD by using
frozen statistics during the forward pass and updating them in a
gradient-free manner, see \Cref{app:coupled}. For self-attention, we
identify the failure mode, but we do not propose a fix here and leave it
to future work.

%% file: s4_experiments.tex
\hypertarget{sec:experiments}{%
\section{Experiments}\label{sec:experiments}}

We evaluate the derivative computation strategies defined in
\Cref{sec:method}: AD, FD and eFD across four PDEs: the Poisson equation
in 2D (Poisson 2D), the 2D heat equation with complex geometry
(Heat2D-CG), the viscous Burgers equation with shock-forming conditions
(Burgers 1D), and finally the 4th-order static Germain--Lagrange
plate-bending equation (Plate).

\hypertarget{sec:setup}{%
\subsection{Experimental setup}\label{sec:setup}}

All experiments use a fully connected MLP with 4 hidden layers and
\(\tanh\) activation. The hidden layer width is 256 for Heat2D-CG and
128 for the other problems. Training uses the Adam optimizer with
learning rate \(10^{-3}\) and a cosine annealing schedule. The loss is

\[
\mathcal{L} = \mathcal{L}_{\mathrm{PDE}} + \lambda_{\mathrm{IC}} \mathcal{L}_{\mathrm{IC}} + \lambda_{\mathrm{BC}} \mathcal{L}_{\mathrm{BC}},
\]

where \(\mathcal{L}_{\mathrm{PDE}}\) is the physics-informed loss and
\(\mathcal{L}_{\mathrm{IC}}\) and \(\mathcal{L}_{\mathrm{BC}}\) are the
initial- and boundary-condition losses, respectively. Loss terms are
weighted with: \(\lambda_{\mathrm{BC}} = 1000\) for Poisson and
\(\lambda_{\mathrm{IC}} = \lambda_{\mathrm{BC}} = 10\) for the other
problems.

Collocation points are resampled uniformly at each iteration with batch
size 8192. One iteration is one optimizer step on a freshly sampled
batch; we do not use a fixed dataset, so there is no notion of an epoch.
The batch size is defined as the number of independent collocation
points; for FD methods the batch dimension of the tensor is actually a
few times larger, see \Cref{tab:sampling} for details. For FD-based
methods, the step size \(\varepsilon\) is recalibrated every 4000
iterations using the empirical procedure of \Cref{sec:calibration}. The
calibration of \(\varepsilon\) (Algorithm 2) uses an independent sample
of \(2^{8+d}\) Sobol points, where \(d\) is the dimensionality of the
input. Each configuration is trained for 20k iterations with multiple
random seeds and metrics are reported as mean \(\pm\) 95\% confidence
interval of the mean (CI; Student \(t\) over seeds). Heat2D-CG
additionally uses an input normalization layer that rescales each input
dimension to \([-1, 1]\); this is essential because the domain
(\(x \in [-8, 8]\), \(y \in [-12, 12]\), \(t \in [0, 3]\)) is far from
the ranges of order one for which a neural network is initialized.
Experiments are run mainly on an NVIDIA GeForce RTX 4070 (12 GB) using
PyTorch \citep{paszke2019pytorch} in FP32 precision, except for the
4th-order Germain--Lagrange plate equation, which is trained in FP64
(\Cref{sec:plate}).

\begin{table}[!ht]
\centering\small
\caption{\label{tab:sampling}\textbf{Summary of batch sizes.} Note the size of the largest single forward pass of the step (dimension 0 of the input tensor). Each loss term is evaluated in a separate pass. The largest one is the PDE residual, where FD and eFD evaluate all stencil points in one batch (5, 7 or 13 per collocation point), while AD evaluates the collocation points alone and obtains the derivatives from the autograd graph of that pass (\Cref{fig:benchmark} compares the resulting memory). $^\dagger$Even though the 13-point biharmonic stencil seems massive (see the derivation in \Cref{app:stencils}), we shall see later, it performs faster and uses less memory than AD on the same problem.}
\begin{tabular}{@{}llrrrrlr@{}}
\toprule
Problem & Inputs & PDE points & \multicolumn{3}{c}{Largest forward pass} & BC points & IC points \\
\cmidrule(lr){4-6}
 & & & AD & FD & eFD & & \\
\midrule
Poisson 2D & $(x, y)$ & 8192 & 8192 & 40\,960 & --- & \begin{tabular}[t]{@{}l@{}}8192\\[-1pt]{\footnotesize 4096 rectangle, 1024 per hole}\end{tabular} & --- \\
\addlinespace[2pt]
Burgers 1D & $(x, t)$ & 8192 & 8192 & 40\,960 & 57\,344 & \begin{tabular}[t]{@{}l@{}}8192\\[-1pt]{\footnotesize 4096 per side}\end{tabular} & 8192 \\
\addlinespace[2pt]
Heat2D-CG & $(x, y, t)$ & 8192 & 8192 & 57\,344 & 57\,344 & \begin{tabular}[t]{@{}l@{}}8192\\[-1pt]{\footnotesize in proportion to perimeter}\end{tabular} & 8192 \\
\addlinespace[2pt]
Plate & $(x, y)$ & 8192 & 8192 & 106\,496$^\dagger$ & 106\,496$^\dagger$ & \begin{tabular}[t]{@{}l@{}}8192\\[-1pt]{\footnotesize 2048 per edge}\end{tabular} & --- \\
\bottomrule
\end{tabular}
\end{table}

\hypertarget{sec:eps}{%
\subsection{\texorpdfstring{Optimal \(\varepsilon\)
selection}{Optimal \textbackslash varepsilon selection}}\label{sec:eps}}

\begin{figure}
\hypertarget{fig:vcurves}{%
\centering
\includegraphics[width=1\textwidth,height=\textheight]{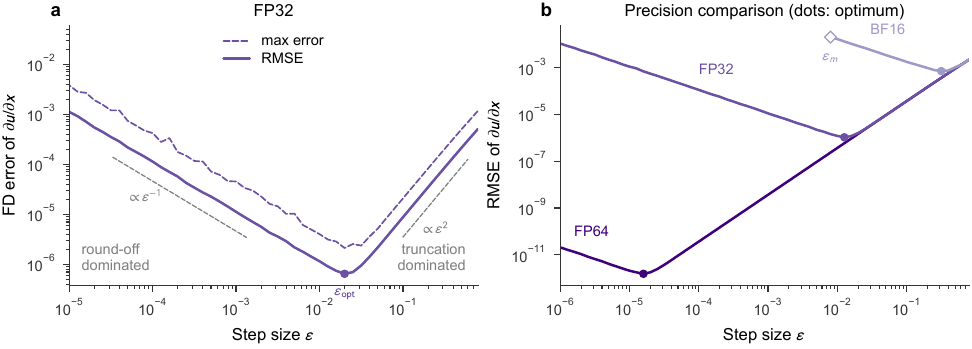}
\caption{Error of the central finite-difference approximation of the
first derivative as a function of the step size \(\varepsilon\). (a)
RMSE and maximum error in FP32: machine round-off dominates for small
\(\varepsilon\), Taylor truncation for large \(\varepsilon\); note the
similar V-shape in Figure 3 of \citet{Baydin2015Automatic}. (b) RMSE for
three floating-point formats; dots mark the empirical optima. BF16, a
format popular in deep learning, is rather impractical for
PINNs.}\label{fig:vcurves}
}
\end{figure}

We validate the theoretical error analysis from \Cref{sec:error-model}
by empirically measuring the finite-difference error as a function of
\(\varepsilon\) on a randomly initialized MLP. For each candidate step
size, we compute the discrepancy between the finite-difference
approximation and the AD-computed reference on a sample of \(2^{8+d}\)
collocation points, i.e.~2048 for the heat equation, and 1024 for the
rest.

\Cref{fig:vcurves}a shows the characteristic V-shaped curve of
finite-difference error as a function of step size \(\varepsilon\).
\Cref{fig:vcurves}b compares the error landscape across three
floating-point formats: FP64, FP32, and BF16. The locations of the
minima follow the theoretical scaling
\(\varepsilon_\mathrm{opt}^{(1)} \sim \varepsilon_m^{1/3}\) of
\Cref{fig:format-eps}a. A practical observation is that BF16 has no
usable window, its optimum sits at \(\varepsilon \approx 0.3\), a step
comparable to the domain itself. Half-precision formats are therefore
unsuitable for finite differences in PINNs, and we do not consider them
further.

\hypertarget{sec:poisson}{%
\subsection{Poisson 2D on a domain with holes}\label{sec:poisson}}

The first problem is a Poisson 2D equation on a rectangular domain with
four circular holes from the PINNacle benchmark \citep{hao2024pinnacle}:

\[
\Delta u(x, y) = 0, \quad (x, y) \in \Omega,
\] with Dirichlet boundary conditions on the outer rectangle and inner
circles. Physically this is the steady temperature of a plate held at a
fixed temperature at its edges and cooled by four pipes. This problem is
stationary, has no initial condition, and involves only 2nd-order
derivatives. eFD and FD therefore coincide on this problem (both use
\(\varepsilon_\mathrm{opt}^{(2)}\)).

\begin{table}[!ht]
\centering\small
\caption{\textbf{Poisson 2D.} Mean \(\pm\) 95\% CI (26 seeds). eFD is
omitted because on a purely second-order PDE it is equivalent to FD.
\label{tab:poisson}}
\begin{tabular}{lllll}
\toprule
Method & L2 Relative (\%) & Max Error (\(\times 10^{-2}\)) & Time (min)
& GPU memory (GB)\tabularnewline
\midrule
AD & 1.41 \(\pm\) 0.13 & 3.22 \(\pm\) 0.34 & 3.3 & 0.27\tabularnewline
FD (= eFD) & 1.40 \(\pm\) 0.16 & 3.26 \(\pm\) 0.56 & 2.1 &
0.17\tabularnewline
\bottomrule
\end{tabular}
\end{table}

\begin{figure}
\hypertarget{fig:poisson}{%
\centering
\includegraphics[width=1\textwidth,height=\textheight]{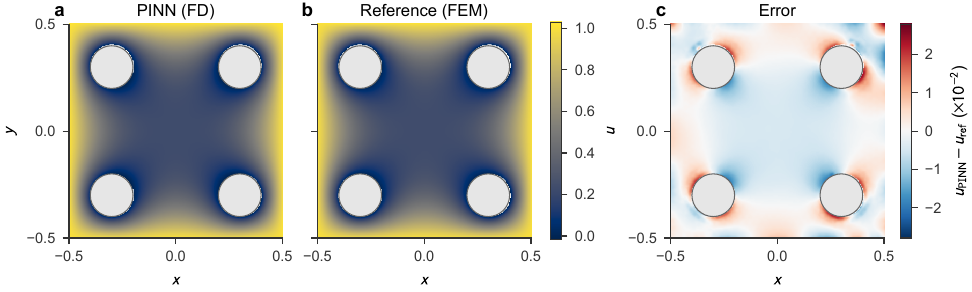}
\caption{Poisson 2D on a square with four circular holes, PINN trained
with FD at the calibrated step: (a) prediction, (b) reference solution
from a high-resolution finite-element solver, (c) pointwise error,
i.e.~raw difference between our solution and the
reference.}\label{fig:poisson}
}
\end{figure}

AD and FD achieve the same accuracy on the Poisson 2D problem
(\Cref{tab:poisson}), while FD is 37\% faster and uses 37\% less GPU
memory; \Cref{fig:poisson} shows an FD solution.

\hypertarget{sec:heat}{%
\subsection{Heat2D-CG (PINNacle complex geometry)}\label{sec:heat}}

The second problem is the 2D heat equation with complex geometry
(Heat2D-CG) from PINNacle \citep{hao2024pinnacle}:

\[
\frac{\partial u}{\partial t} - \Delta u = 0, \quad (x, y) \in \Omega, \quad t \in [0, 3],
\] where \(\Omega \subset [-8, 8] \times [-12, 12]\) is a rectangle with
17 circular holes representing the tubes of a heat exchanger; transient
conduction around such tubes is a routine thermal-design problem. The
initial condition is \(u(x, y, 0) = 0\). The boundary conditions are of
Robin type, \[
\nabla u \cdot \mathbf{n} = g - q \cdot u,
\] with parameters \(g\) and \(q\) varying across boundary segments
(large circles: \(g = 5\), \(q = 1\); small circles: \(g = 1\),
\(q = 1\); outer rectangle: \(g = 0.1\), \(q = 1\)). The reference
solution was obtained from COMSOL on an unstructured mesh with 13,312
nodes and 31 time steps by \citet{hao2024pinnacle}. This problem
introduces two additional challenges for FD: the domain is
three-dimensional (\(x\), \(y\), \(t\)), and the Robin boundary
conditions involve normal derivatives \(\nabla u \cdot \mathbf{n}\),
where the normal direction varies along the curved boundaries.

\begin{table}[!ht]
\centering\small
\caption{\textbf{Heat2D-CG.} Mean \(\pm\) 95\% CI (26 seeds).
\label{tab:heat}}
\begin{tabular}{lllll}
\toprule
Method & L2 Relative (\%) & Max Error (\(\times 10^{-1}\)) & Time (min)
& GPU memory (GB)\tabularnewline
\midrule
AD & 2.56 \(\pm\) 0.13 & 5.71 \(\pm\) 0.10 & 9.0 & 0.63\tabularnewline
FD & 2.66 \(\pm\) 0.16 & 5.638 \(\pm\) 0.075 & 9.3 & 0.54\tabularnewline
eFD & 2.60 \(\pm\) 0.13 & 5.773 \(\pm\) 0.097 & 9.3 &
0.54\tabularnewline
\bottomrule
\end{tabular}
\end{table}

\begin{figure}
\hypertarget{fig:heat}{%
\centering
\includegraphics[width=1\textwidth,height=\textheight]{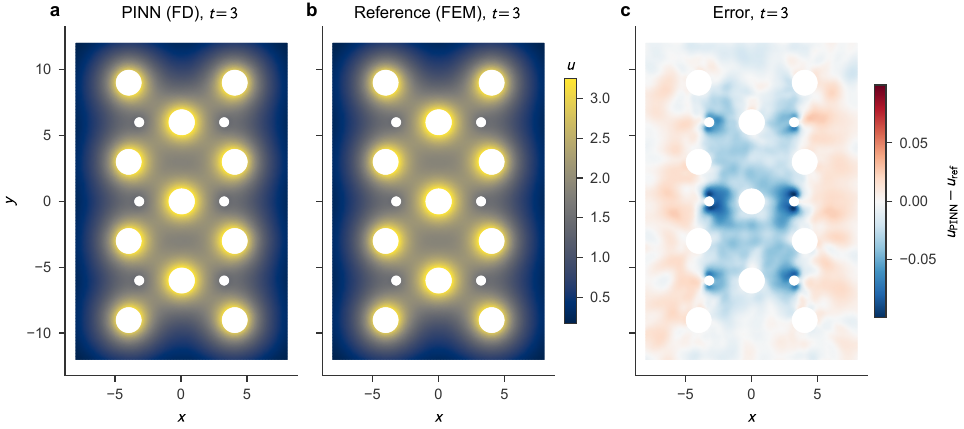}
\caption{Heat2D-CG, PINN trained with FD at the calibrated step, at the
final time \(t = 3\): (a) prediction, (b) finite-element reference from
\citep{hao2024pinnacle}, (c) pointwise error, i.e.~raw difference
between our solution and the reference.}\label{fig:heat}
}
\end{figure}

All three methods are statistically indistinguishable (\Cref{tab:heat}).
Our AD baseline is better than the original PINNacle vanilla PINN
(3.64\%) and close to the best method reported in their benchmark (LAAF,
2.39\%), confirming that our experimental configuration is a solid
implementation. \Cref{fig:heat} shows an FD solution at the final time.

\hypertarget{sec:burgers}{%
\subsection{1D viscous Burgers equation}\label{sec:burgers}}

We consider the one-dimensional viscous Burgers equation, a simplified
model of fluid dynamics describing nonlinear wave steepening moderated
by viscous diffusion:

\[
\frac{\partial u}{\partial t} + u \frac{\partial u}{\partial x} = \nu \frac{\partial^2 u}{\partial x^2}, \quad x \in [-1, 1], \quad t \in [0, 1],
\]

with viscosity \(\nu = 0.01/\pi\), initial condition
\(u(x, 0) = -\sin(\pi x)\), and homogeneous Dirichlet boundary
conditions \(u(\pm 1, t) = 0\). This nonlinear problem develops a sharp
shock front around \(t \approx 0.3\) and is a standard stress test for
PINNs \citep{raissi2019pinn, hao2024pinnacle, Kavousanakis2025Flow}. The
PDE combines 1st-order (\(u_t\), \(u_x\)) and 2nd-order (\(u_{xx}\))
derivatives, directly exercising the mixed-order step-size strategies.

The reference solution is the exact solution obtained with the
Cole--Hopf transformation \citep{Hopf1950Partial, Cole1951Quasilinear},
evaluated by Gauss--Hermite quadrature \citep{Basdevant1986Spectral} on
the grid of the PINNacle benchmark \citep{hao2024pinnacle} (101 points
in \(x\), 11 time steps). We replaced the finite-element reference
shipped with PINNacle after finding that it differs from the exact
solution by 1.33\% in relative L2, but we present it in the figures for
completeness. It is also an interesting observation that
well-established finite-element solvers, if not applied carefully, may
lead to a few times higher errors than a well-calibrated PINN. That also
opens a discussion whether using FEM as a reference for comparing PINN
methods is still a valid approach. That is also why, to test a 4th-order
PDE in the following section, we used a problem with a known analytic
solution.

\begin{table}[!ht]
\centering\small
\caption{\textbf{Burgers 1D.} Mean \(\pm\) 95\% CI (26 seeds).
\label{tab:burgers}}
\begin{tabular}{lllll}
\toprule
Method & L2 Relative (\%) & Max Error (\(\times 10^{-2}\)) & Time (min)
& GPU memory (GB)\tabularnewline
\midrule
AD & 0.52 \(\pm\) 0.12 & 4.7 \(\pm\) 1.4 & 2.9 & 0.21\tabularnewline
FD & 0.62 \(\pm\) 0.12 & 6.3 \(\pm\) 1.7 & 2.4 & 0.18\tabularnewline
eFD & 0.52 \(\pm\) 0.11 & 4.9 \(\pm\) 1.5 & 2.6 & 0.22\tabularnewline
\bottomrule
\end{tabular}
\end{table}

\begin{figure}
\hypertarget{fig:burgers}{%
\centering
\includegraphics[width=1\textwidth,height=\textheight]{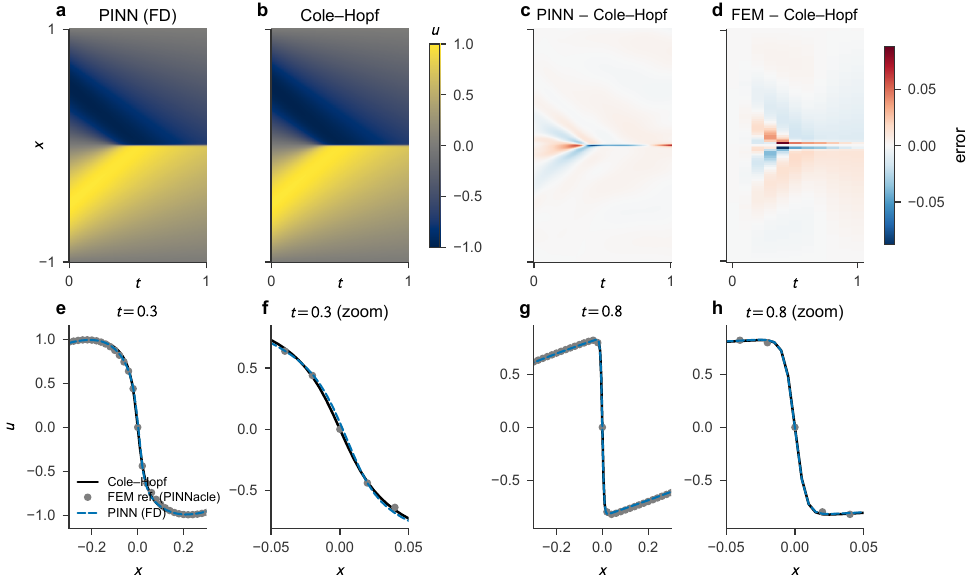}
\caption{Burgers 1D, PINN trained with FD at the calibrated step: (a)
prediction, (b) Cole--Hopf reference, (c) pointwise error, i.e.~raw
difference between our solution and the Cole--Hopf reference, (d)
difference between the finite-element solution of PINNacle
\citep{hao2024pinnacle} and the Cole--Hopf reference. (e--h) Profiles at
\(t = 0.3\) and \(t = 0.8\), each with a zoom on the
shock.}\label{fig:burgers}
}
\end{figure}

AD, FD, and eFD agree within their CIs (\Cref{tab:burgers};
\Cref{fig:burgers} shows an FD solution and the profiles across the
shock), although it is visible that eFD, i.e.~\(\varepsilon\) calibrated
per derivative order, gives a slightly better result than using the same
\(\varepsilon\) for both orders. GPU memory usage and time are
comparable for all methods with no clear winner.

\hypertarget{sec:plate}{%
\subsection{Germain--Lagrange plate (4th order)}\label{sec:plate}}

Finally, to test the method beyond second derivatives we addressed
Kirchhoff--Love bending of a simply supported square plate under a
uniform load, governed by the Germain--Lagrange equation, the basic
model of thin plates in structural mechanics (floor slabs, panels, ship
decks): \[
\Delta^2 w = \frac{\partial^4 w}{\partial x^4} + 2\frac{\partial^4 w}{\partial x^2 \partial y^2} + \frac{\partial^4 w}{\partial y^4} = \frac{q}{D}, \quad (x, y) \in (0, 1)^2, \qquad w = 0, \quad \Delta w = 0 \ \text{on the boundary}.
\]

The reference is the classical Navier double-sine series
\citep{timoshenko1959plates}, truncated at \(m, n \le 99\) (truncation
error below \(10^{-10}\)); the load is normalized so that the maximum
deflection equals one. The problem mixes derivative orders in a
different way than Burgers: the fourth order (including the mixed term)
appears in the PDE, the second order in the boundary condition. FD
evaluates \(\Delta^2 w\) with the standard 13-point stencil
\citep{abramowitz1964handbook, Bjorstad1983Fast}; its weights and
derivation are given in \Cref{app:stencils}.

The network architecture, optimizer, schedule and loss weights are the
same as for the Burgers equation.

\begin{table}[!ht]
\centering\small
\caption{\label{tab:plate}\textbf{Plate (4th order).} Mean $\pm$ 95\% CI (10 seeds).}
\begin{tabular}{@{}ll>{\hspace{1em}}c>{\hspace{1em}}c>{\hspace{1em}}rr@{}}
\toprule
Method & Precision & L2 Relative (\%) & Max Error ($\times 10^{-3}$) & Time (min) & GPU memory (GB) \\
\midrule
AD & FP64 & 0.0477 $\pm$ 0.0071 & 1.72 $\pm$ 0.32 & 133 & 4.51 \\
FD & FP64 & 0.0482 $\pm$ 0.0081 & 1.70 $\pm$ 0.34 & 44 & 0.82 \\
eFD & FP64 & 0.0486 $\pm$ 0.0067 & 1.74 $\pm$ 0.27 & 44 & 0.82 \\
\midrule
AD & FP32 & 0.0523 $\pm$ 0.0069 & 1.78 $\pm$ 0.30 & 25 & 2.26 \\
FD & FP32 & 0.098 $\pm$ 0.043 & 1.56 $\pm$ 0.18 & 3.5 & 0.42 \\
\bottomrule
\end{tabular}
\end{table}

\begin{figure}
\hypertarget{fig:plate-cost}{%
\centering
\includegraphics[width=1\textwidth,height=\textheight]{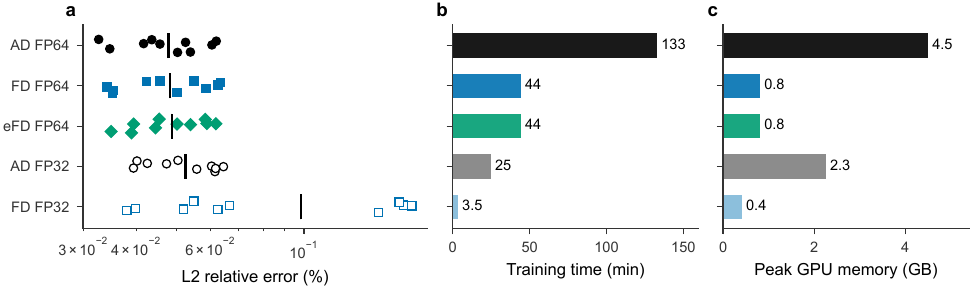}
\caption{Visualization of \Cref{tab:plate}: (a) L2 relative error of
every training run, the bar marks the mean; (b) median training time;
(c) median peak GPU memory. Note the large variance of the L2 relative
error for FD FP32.}\label{fig:plate-cost}
}
\end{figure}

As shown in \Cref{tab:plate} and \Cref{fig:plate-cost}, both
finite-difference variants reproduce the AD results in about a third of
the time and with less than a fifth of the memory. We also found that
the FP32 format (used successfully for the 2nd-order equations) limits
the accuracy of the final solution obtained with FD to about 0.1\%,
about twice the FP64 value. This is likely caused by the fact that
4th-order differences amplify round-off error as \(\varepsilon^{-4}\).
We admit that AD is more robust to the floating-point format and
achieves comparable accuracy in both precisions. However, in double
precision (FP64), the finite-difference methods are a clear winner.

\hypertarget{sec:cost}{%
\subsection{Computational cost}\label{sec:cost}}

\begin{figure}
\hypertarget{fig:benchmark}{%
\centering
\includegraphics[width=1\textwidth,height=\textheight]{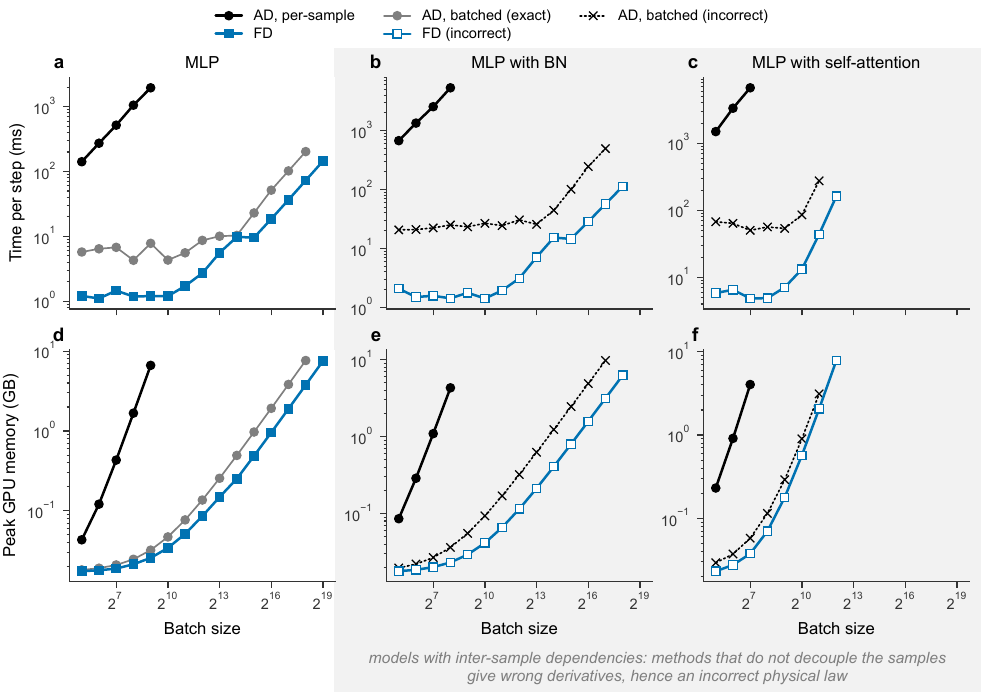}
\caption{Wall-clock time per training step (top) and peak GPU memory
(bottom) against batch size for three MLP variants. For the plain MLP
(a, d) batched AD and FD give correct derivatives and per-sample AD is
not required; it is shown for completeness. This is the setting of all
models used in this paper, and FD beats AD in both speed and memory.
When the MLP contains BN (b, e), both batched methods give wrong results
and per-sample AD is needed. Similarly, for an MLP with self-attention
(c, f) only per-sample AD gives correct derivatives, at the cost of an
impractically small batch size (\(2^7\) against our \(2^{13}\)) and of
training speed.}\label{fig:benchmark}
}
\end{figure}

Apart from reporting the total training time in previous sections, we
measured the isolated time of the PINN training step (Poisson residual
evaluation followed by a regular backpropagation pass through the
network parameters; \Cref{fig:benchmark}). We also tested the maximum
batch size that each method allows before hitting out-of-memory errors.
The plain MLP is the same neural network architecture as used in all
problems. The other two variants contain either batch normalization
layers, precisely PyTorch \passthrough{\lstinline!BatchNorm1d!}, or
self-attention layers, namely PyTorch
\passthrough{\lstinline!MultiheadAttention!}. Interestingly, we observed
that for the last two variants, the only correct method, i.e.~per-sample
AD, hits the 12 GB GPU limit already at batch sizes above 128, which, in
our humble experience, is below what practical applications need.

\hypertarget{sec:interval}{%
\subsection{\texorpdfstring{Sensitivity to the recalibration interval of
\(\varepsilon\)}{Sensitivity to the recalibration interval of \textbackslash varepsilon}}\label{sec:interval}}

\begin{figure}
\hypertarget{fig:ablation}{%
\centering
\includegraphics[width=1\textwidth,height=\textheight]{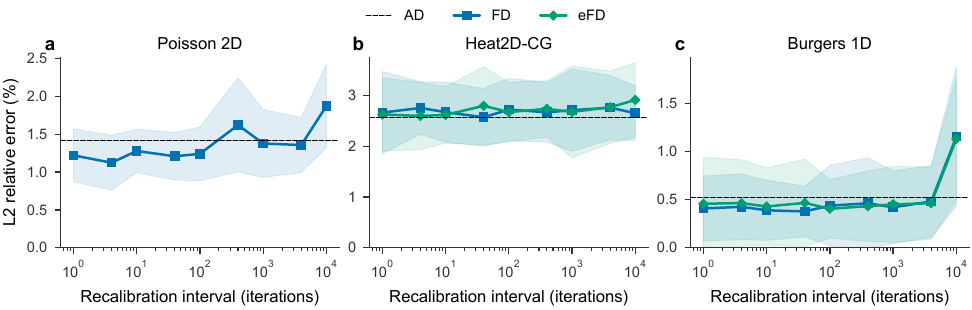}
\caption{Ablation of the \(\varepsilon\) recalibration interval on (a)
Poisson 2D, (b) Heat2D-CG and (c) Burgers 1D: L2 relative error
vs.~interval, mean \(\pm\) 95\% CI (5 seeds); the dashed line is the AD
level. FD and eFD are robust to the interval across four orders of
magnitude, but too infrequent recalibration (a, c) can lead to
substantially higher errors.}\label{fig:ablation}
}
\end{figure}

\Cref{sec:calibration} argued that the optimal step size evolves as the
network's smoothness characteristics change during training and
recommended periodic recalibration. To validate this claim we run each
method at refresh intervals of
\(\{1, 4, 10, 40, 100, 400, 1000, 4000, 10000\}\) iterations on the
three 2nd-order problems (5 seeds each, 20k iterations).
\Cref{fig:ablation} reports the L2 relative error as a function of the
interval.

We observe that all FD methods are robust to the recalibration
frequency. Across all three problems, their L2 error is essentially flat
from interval 1 to interval 4000. Only at interval 10000, i.e.~when
\(\varepsilon\) is calibrated twice during the entire training, the FD
and eFD errors increase above the AD level.

\hypertarget{sec:sfd}{%
\subsection{Bonus: Stochastic Finite Difference}\label{sec:sfd}}

We also explored another FD variant where the step size \(\varepsilon\)
is not fixed, but randomly drawn from some interval. We name it
Stochastic Finite Difference (sFD).

The interval spans the optima of the highest derivative order of the
problem and of the order below it:
\([\varepsilon_\mathrm{opt}^{(1)}, \varepsilon_\mathrm{opt}^{(2)}]\) on
the 2nd-order problems (on Poisson as well, although its PDE contains
only second derivatives) and
\([\varepsilon_\mathrm{opt}^{(3)}, \varepsilon_\mathrm{opt}^{(4)}]\) on
the plate, where the boundary Laplacian samples from the same interval.

sFD performs substantially worse than the other methods on the
time-dependent problems, is indistinguishable from them on the plate,
and works surprisingly well on the Poisson equation (\Cref{tab:sfd},
\Cref{sec:poisson}). We suppose that changing the effective length scale
visible to the gradient prevents a solution collapse. We present it here
separately as an interesting concept that needs further investigation
rather than a competitive method.

\begin{table}[!ht]
\centering\small
\caption{\textbf{Exploratory method: stochastic step sizes.} L2 relative
error (\%), mean \(\pm\) 95\% CI. sFD draws the step log-uniformly
between the calibrated optima of the highest derivative order and the
order below it: \([\varepsilon^{(1)}, \varepsilon^{(2)}]\) on the
second-order problems (26 seeds),
\([\varepsilon^{(3)}, \varepsilon^{(4)}]\) on the plate (10 seeds).
Bold: best, with non-overlapping CIs. \label{tab:sfd}}
\begin{tabular}{llll}
\toprule
Problem & AD & FD & sFD\tabularnewline
\midrule
Poisson 2D & 1.41 \(\pm\) 0.13 & 1.40 \(\pm\) 0.16 & 1.11 \(\pm\)
0.16\tabularnewline
Heat2D-CG & \textbf{2.56 \(\pm\) 0.13} & \textbf{2.66 \(\pm\) 0.16} &
4.65 \(\pm\) 0.92\tabularnewline
Burgers 1D & \textbf{0.52 \(\pm\) 0.12} & \textbf{0.62 \(\pm\) 0.12} &
3.82 \(\pm\) 0.43\tabularnewline
Plate (4th order) & 0.0477 \(\pm\) 0.0071 & 0.0482 \(\pm\) 0.0081 &
0.0495 \(\pm\) 0.0092\tabularnewline
\bottomrule
\end{tabular}
\end{table}

%% file: s5_discussion.tex
\hypertarget{sec:conclusions}{%
\section{Conclusions}\label{sec:conclusions}}

Finite differences are a viable alternative to automatic differentiation
in PINNs, provided that the step size \(\varepsilon\) is calibrated
carefully. In particular, they come to the rescue when high-order PDEs
lead to a catastrophic growth of the memory needed to accommodate the
exploding AD graph: on the plate, FD trains 3× faster and needs about 6×
less GPU memory than AD (\Cref{sec:plate}), and the microbenchmark of
\Cref{sec:cost} shows that the gap widens with the batch size.

Step-size calibration, which is often given little attention in
finite-difference approaches, turns out to be critical. Contrary to
naive intuition, a smaller step size does not always lead to better
accuracy. As shown in \Cref{sec:error-model}, the optimal step balances
round-off and truncation errors, and grows with the derivative order.
The simple method proposed in \Cref{sec:calibration} finds the optimal
step size in seconds. Moreover, such a calibration should be repeated a
few times during the training, but too frequent recalibration brings
little to no improvement, as shown in \Cref{sec:interval}.

For PDEs that mix derivative orders, a few optimal step sizes exist. But
in practice the single geometric-mean step (FD) and the order-specific
steps (eFD) perform alike (\Cref{sec:heat,sec:burgers,sec:plate}).

Numerical precision determines the range of viable step sizes
\(\varepsilon\). As shown in \Cref{sec:eps}, BF16 offers no practical
operating range: its optimal step size is comparable to the domain size,
rendering it unsuitable for finite differences in PINNs. FP32 provides
sufficient accuracy for second-order derivatives, as demonstrated in
\Cref{sec:poisson,sec:heat,sec:burgers}. For fourth-order derivatives
(\Cref{sec:plate}), however, FP32 limits the accuracy of FD, because
round-off errors are scaled proportionally to \(\varepsilon^{-4}\). In
that case FP64 is preferred. Automatic differentiation is less sensitive
to numerical precision but also profits from double precision.

As illustrated with the Burgers equation in \Cref{sec:burgers}, FEM
solutions are not always a good reference. Problems with known
nontrivial analytical solutions are probably better suited as test beds.

Finally, as shown in \Cref{sec:persample} and \Cref{app:coupled}, the
common \passthrough{\lstinline!grad\_outputs=ones!} idiom is silently
wrong in neural architectures with inter-sample dependencies --- and we
believe that addressing this limitation is crucial for opening the stage
for powerful architectures with self-attention layers. For batch
normalization a layer that uses only its running statistics removes the
coupling (\Cref{app:coupled}); the fix for self-attention, as well as
the stochastic step-size variant, we leave for future work.

%% file: s6_declarations.tex
\hypertarget{data-availability}{%
\section*{Data availability}\label{data-availability}}
\addcontentsline{toc}{section}{Data availability}

The reference solutions used in this study are obtained from the
PINNacle benchmark suite \citep{hao2024pinnacle}: a finite-element
solution for the Poisson 2D problem on a domain with four circular holes
and the Heat2D-CG dataset (COMSOL), together with the evaluation grid of
the 1D viscous Burgers problem, on which we evaluate the exact
Cole--Hopf solution (\Cref{sec:burgers}); the plate reference is the
Navier series. All numerical training outputs reported in this
manuscript --- per-seed L2 relative errors, SMAPE, maximum errors,
training wall-clock times, calibrated step sizes \(\varepsilon\), and
the speed/memory microbenchmark measurements --- are stored as CSV files
together with the arrays needed to regenerate the figures in
\passthrough{\lstinline!.npz!} format. The data behind every table and
figure will be made available upon publication in the reproduction
repository, https://github.com/mjmikulski/finite-pinn-repro, archived on
Zenodo (DOI: 10.5281/zenodo.23709828).

\hypertarget{code-availability-and-simulation-software}{%
\section*{Code availability and simulation
software}\label{code-availability-and-simulation-software}}
\addcontentsline{toc}{section}{Code availability and simulation
software}

All custom code used to generate the results in this manuscript ---
PyTorch implementations of the derivative-computation strategies (AD,
FD, eFD), training scripts for the four benchmark PDEs, the empirical
\(\varepsilon\) calibration routine, the speed/memory microbenchmark ---
will be made available upon publication in the same repository, together
with a demonstration of the batched-AD pitfall of \Cref{app:coupled} and
tests that check the main claims of the paper on a CPU\ifreview; a
snapshot is provided to the reviewers as Supporting Information\fi.
Experiments were run with Python 3.11 or 3.12, and PyTorch 2.11.0 on a
single NVIDIA RTX 4070 GPU. Some experimental runs (not used in speed
benchmarks) were executed on a secondary GPU (NVIDIA RTX 2070S). No
proprietary software was used in the study and everyone should be able
to reproduce the results, which we warmly encourage.

\hypertarget{funding}{%
\section*{Funding}\label{funding}}
\addcontentsline{toc}{section}{Funding}

This work did not receive specific external funding.

\hypertarget{use-of-generative-ai-tools}{%
\section*{Use of generative AI tools}\label{use-of-generative-ai-tools}}
\addcontentsline{toc}{section}{Use of generative AI tools}

The authors used a generative AI coding assistant (Anthropic's Claude,
accessed via the Claude Code command-line interface) throughout the
preparation of this manuscript. The assistant was used for: (i)
refactoring and writing portions of the research codebase, including the
figure-generation scripts and parts of the benchmark and ablation
pipelines; (ii) drafting, restructuring, and copy-editing prose in
several sections of the manuscript; (iii) literature search and
summarization; and (iv) producing summary tables and analysis of
experimental results from CSV outputs. All scientific contributions ---
the research questions, methodology, experimental design, interpretation
of results and final wording of every claim --- were conceived, verified
and approved by the human authors, who take full responsibility for the
content. The AI assistant is not credited as an author.

\hypertarget{author-contributions-credit}{%
\section*{Author contributions
(CRediT)}\label{author-contributions-credit}}
\addcontentsline{toc}{section}{Author contributions (CRediT)}

\textbf{M.J.M.:} conceptualization, methodology, software,
investigation, formal analysis, visualization, writing --- original
draft. \textbf{T.U.:} supervision, methodology, writing --- review \&
editing.

\hypertarget{competing-interests}{%
\section*{Competing interests}\label{competing-interests}}
\addcontentsline{toc}{section}{Competing interests}

M.J.M. is employed by NVIDIA. The work reported in this manuscript was
conducted independently, outside the author's employment
responsibilities, using private computational resources. NVIDIA had no
role in the study design, experiments, analysis, interpretation,
manuscript preparation, or decision to submit the work. The remaining
author declares no competing interests.

%% file: s7_appendix.tex
\hypertarget{app:stencils}{%
\section{Finite-difference stencils in higher dimensions and at higher
order}\label{app:stencils}}

For a neural network \(u_\theta : \mathbb{R}^d \to \mathbb{R}\) and a
batch of collocation points \(X \in \mathbb{R}^{B \times d}\), the
central-difference stencil for the Laplacian is

\[
\Delta u_\theta(x) \approx \frac{1}{\varepsilon^2} \sum_{k=1}^{d} \left[ u_\theta(x + \varepsilon e_k) - 2 u_\theta(x) + u_\theta(x - \varepsilon e_k) \right],
\]

where \(e_k\) is the \(k\)-th unit vector in input space. The total
number of function evaluations is \(2d + 1\) per collocation point, and
all of them can be packed into a single forward pass through the network
by concatenating the perturbed batches along the batch dimension:

\begin{lstlisting}[language=Python]
def laplacian_fd(model, X, eps):
    # X: (B, d), model: R^d -> R, eps: scalar step size
    B, d = X.shape
    perturbations = [X]  # center
    for k in range(d):
        shift = torch.zeros(d, device=X.device)
        shift[k] = eps
        perturbations.append(X + shift)
        perturbations.append(X - shift)
    X_all = torch.cat(perturbations, dim=0)  # (B * (2d+1), d)
    u_all = model(X_all).view(2 * d + 1, B)
    center = u_all[0]
    plus  = u_all[1::2]  # (d, B)
    minus = u_all[2::2]  # (d, B)
    lap = (plus.sum(0) + minus.sum(0) - 2 * d * center) / (eps ** 2)
    return lap
\end{lstlisting}

The one-dimensional central stencils of 2nd-order accuracy for the third
and fourth derivatives, used for the calibration of
\(\varepsilon^{(3)}\) and \(\varepsilon^{(4)}\) on the plate
(\Cref{sec:plate}), are

\[
D_\varepsilon^{(3)} f(x) = \frac{-f(x - 2\varepsilon) + 2 f(x - \varepsilon) - 2 f(x + \varepsilon) + f(x + 2\varepsilon)}{2\varepsilon^3},
\] \[
D_\varepsilon^{(4)} f(x) = \frac{f(x - 2\varepsilon) - 4 f(x - \varepsilon) + 6 f(x) - 4 f(x + \varepsilon) + f(x + 2\varepsilon)}{\varepsilon^4}.
\]

\hypertarget{app:biharmonic}{%
\subsection{The 13-point biharmonic stencil}\label{app:biharmonic}}

\begin{figure}
\hypertarget{fig:stencil}{%
\centering
\includegraphics[width=1\textwidth,height=\textheight]{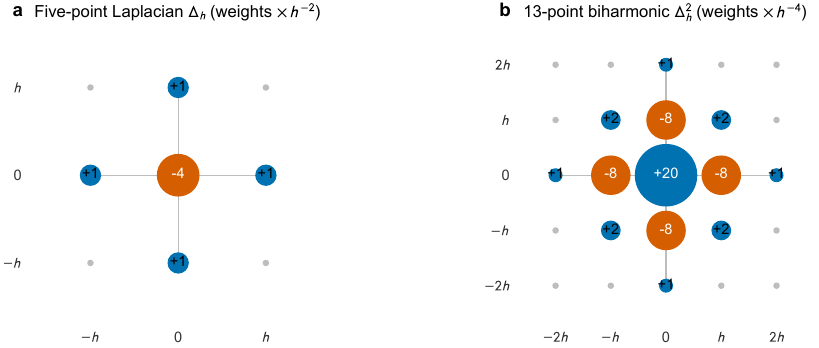}
\caption{The five-point Laplacian (a) and the 13-point biharmonic
stencil obtained by applying it twice (b). Circle area is proportional
to the magnitude of the weight, color to its sign; the weights are in
units of \(h^{-2}\) and \(h^{-4}\), respectively.}\label{fig:stencil}
}
\end{figure}

On a square grid of spacing \(h\) the five-point Laplacian reads

\[
\Delta_h u_{i,j} = \frac{u_{i+1,j} + u_{i-1,j} + u_{i,j+1} + u_{i,j-1} - 4 u_{i,j}}{h^2},
\] with truncation error \(O(h^2)\). The biharmonic operator
\(\Delta^2 = \Delta\Delta\) is discretized by applying this stencil
twice, \(\Delta_h^2 u = \Delta_h(\Delta_h u)\). Substituting the
five-point formula into itself collects, for each neighbor, the products
of the weights along all two-step paths from the center: the center
receives \((-4)^2 + 4 \cdot 1 = 20\), each of the four axis neighbors
\(2 \cdot (-4) \cdot 1 = -8\), each diagonal neighbor \(1 + 1 = 2\) (two
paths), and each point at distance \(2h\) along an axis
\(1 \cdot 1 = 1\). Hence

\[
\Delta_h^2 u_{i,j} = \frac{1}{h^4} \Big[ 20\, u_{i,j} - 8 \big( u_{i \pm 1, j} + u_{i, j \pm 1} \big) + 2 \big( u_{i \pm 1, j \pm 1} \big) + \big( u_{i \pm 2, j} + u_{i, j \pm 2} \big) \Big] + O(h^2),
\] where each parenthesis sums over the indicated sign combinations
(\Cref{fig:stencil}). The weights sum to zero, as they must for a
derivative operator, and the stencil contains the mixed term
\(\partial^4 u / \partial x^2 \partial y^2\) through the diagonal
points: it is the smallest central stencil of 2nd-order accuracy for
\(\Delta^2\), and the classical one for the plate equation
\citetext{\citealp[formula
25.3.32]{abramowitz1964handbook}; \citealp{Bjorstad1983Fast}}. In
training, the 13 shifted copies of the collocation batch are evaluated
in one forward pass (\Cref{tab:sampling}), and the boundary Laplacian
uses the five-point stencil above.

\hypertarget{app:error}{%
\section{Derivation of the error model}\label{app:error}}

\textbf{First derivative.} Expanding \(f(x \pm \varepsilon)\) in Taylor
series about \(x\), \[
f(x \pm \varepsilon) = f(x) \pm \varepsilon f'(x) + \frac{\varepsilon^2}{2}f''(x) \pm \frac{\varepsilon^3}{6}f'''(x) + O(\varepsilon^4),
\] subtracting the two expansions and dividing by \(2\varepsilon\)
yields \[
D_\varepsilon^{(1)} f(x) = f'(x) + \frac{\varepsilon^2}{6}f'''(x) + O(\varepsilon^3),
\] i.e.~a truncation error
\(E_{\mathrm{trunc}}^{(1)}(\varepsilon) = \frac{\varepsilon^2}{6}|f'''(x)| + O(\varepsilon^3)\).
For the round-off error, each evaluation \(f(x \pm \varepsilon)\) incurs
a relative error of order \(\varepsilon_m\) (machine epsilon),
corresponding to an absolute error of approximately
\(\varepsilon_m |f|\). The subtraction of two nearly equal quantities
and the division by \(2\varepsilon\) amplify this error; the precise
factor depends on implementation details and cancellation, so we write
\[
E_{\mathrm{round}}^{(1)}(\varepsilon) = \kappa_1 \frac{\varepsilon_m |f(x)|}{\varepsilon},
\] with an unknown constant \(\kappa_1\). Dropping \(O(\varepsilon^3)\)
and setting \(dE^{(1)}/d\varepsilon = 0\) for the sum of the two terms,
\[
\frac{\varepsilon}{3}|f'''(x)| = \kappa_1 \frac{\varepsilon_m |f(x)|}{\varepsilon^2}
\quad \Longrightarrow \quad
\varepsilon_\mathrm{opt}^{(1)} = \left( \frac{3 \kappa_1 \varepsilon_m |f(x)|}{|f'''(x)|} \right)^{1/3} = C_1 \cdot \varepsilon_m^{1/3},
\] with \(C_1 = (3 \kappa_1)^{1/3} \left( |f|/|f'''| \right)^{1/3}\).

\textbf{Second derivative.} Adding the Taylor expansions, \[
f(x + \varepsilon) + f(x - \varepsilon) = 2f(x) + \varepsilon^2 f''(x) + \frac{\varepsilon^4}{12}f''''(x) + O(\varepsilon^6),
\] from which \[
D_\varepsilon^{(2)} f = f''(x) + \frac{\varepsilon^2}{12}f''''(x) + O(\varepsilon^4),
\]
i.e.~\(E_{\mathrm{trunc}}^{(2)} = \frac{\varepsilon^2}{12}|f''''(x)| + O(\varepsilon^4)\).
The round-off error, amplified by the division by \(\varepsilon^2\), is
\(E_{\mathrm{round}}^{(2)} = \kappa_2\, \varepsilon_m |f(x)| / \varepsilon^2\).
Minimizing the total error, \[
\frac{\varepsilon}{6}|f''''(x)| = \frac{2\kappa_2 \varepsilon_m |f(x)|}{\varepsilon^3}
\quad \Longrightarrow \quad
\varepsilon_\mathrm{opt}^{(2)} = \left( \frac{12 \kappa_2 \varepsilon_m |f(x)|}{|f''''(x)|} \right)^{1/4} = C_2 \cdot \varepsilon_m^{1/4}.
\]

\textbf{General order.} For a central stencil of order \(k\) with
2nd-order accuracy the truncation error is \(a_k \varepsilon^2\) with
\(a_k \propto |f^{(k+2)}(x)|\) and the round-off error
\(b_k \varepsilon_m \varepsilon^{-k}\) with \(b_k \propto |f(x)|\); the
same minimization gives
\(\varepsilon_\mathrm{opt}^{(k)} = (k b_k / 2 a_k)^{1/(k+2)}\, \varepsilon_m^{1/(k+2)}\),
the formula of \Cref{sec:error-model}.

\hypertarget{app:coupled}{%
\section{Derivatives in architectures with inter-sample
dependencies}\label{app:coupled}}

\begin{figure}
\hypertarget{fig:architectures}{%
\centering
\includegraphics[width=1\textwidth,height=\textheight]{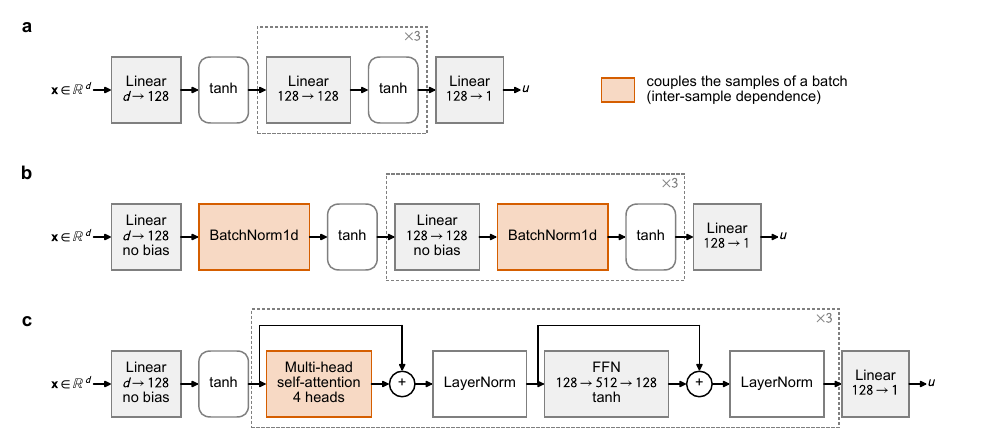}
\caption{The three architectures of this appendix: (a) plain MLP, (b)
MLP with BN, (c) MLP with self-attention. Shaded: the layers that couple
the samples of a batch. The self-attention blocks treat the batch of
collocation points as a sequence of tokens, so that each output depends
on all \(B\) collocation points through the attention
weights.}\label{fig:architectures}
}
\end{figure}

This appendix gives the numerical evidence behind the
\passthrough{\lstinline!grad\_outputs=ones!} failure mode of
\Cref{sec:persample}, for the three architectures of
\Cref{fig:architectures}, and shows that for BN the coupling can be
removed at the layer level (Algorithm 3).

The networks are pretrained for 2000 iterations on the smooth target
\(u(x, y) = \sin x \cdot \cos y\) (Adam, learning rate \(10^{-3}\),
batch size 1024) rather than on a PDE: training these models on a PDE
residual requires precisely the per-sample derivative whose computation
is examined here, so we use a synthetic target instead, to focus purely
on the derivative problem. Pretraining also brings the output magnitudes
of the architectures into a comparable range and, for BN, settles the
running statistics.

For each architecture we compare three ways of computing the Laplacian
\(\Delta u\) of the network output on a batch of \(B = 1024\)
collocation points: the \passthrough{\lstinline!grad\_outputs=ones!}
idiom applied twice (batched AD); the correct sequential loop
(per-sample AD, the reference); and the five-point FD stencil with
\(\varepsilon\) calibrated for each network by Algorithm 2. For the
plain MLP and MLP with BN, all five stencil copies are concatenated into
one batch of \(5B\) points; for the self-attention model each stencil
copy is a separate sequence of \(B\) tokens. All computations are in
FP64. \Cref{tab:bn} reports the relative L2 discrepancy
\(\|\cdot - \mathrm{AD}_\mathrm{ps}\|_2 / \|\mathrm{AD}_\mathrm{ps}\|_2\)
from the per-sample reference \(\mathrm{AD}_\mathrm{ps}\) (the L2
relative error of the main text, with the per-sample derivative in place
of the exact solution), median over 5 random initializations.

\begin{table}[!ht]
\centering\small
\caption{\label{tab:bn}\textbf{Laplacian estimates on coupled architectures.} Relative L2 discrepancy $\|\text{batched AD} - \mathrm{AD}_\mathrm{ps}\|_2 / \|\mathrm{AD}_\mathrm{ps}\|_2$ and $\|\text{FD} - \mathrm{AD}_\mathrm{ps}\|_2 / \|\mathrm{AD}_\mathrm{ps}\|_2$ from the per-sample AD reference $\mathrm{AD}_\mathrm{ps}$, FD at the calibrated step $\varepsilon$; median over 5 random initializations, $B = 1024$, FP64 (script in the accompanying code repository). For self-attention the calibration finds no step (see text), so FD is not reported.}
\begin{tabular}{@{}llrr@{}}
\toprule
Architecture & $\varepsilon$ (calibrated) & \multicolumn{2}{c}{relative L2 discrepancy} \\
\cmidrule(lr){3-4}
 & & batched AD & FD \\
\midrule
plain MLP & $1.8 \times 10^{-4}$ & $0$ & $2 \times 10^{-6}\,\%$ \\
MLP with BN & $4.5 \times 10^{-3}$ & $100\,\%$ & $0.5\,\%$ \\
MLP with decoupled BN (Algorithm 3) & $4.0 \times 10^{-5}$ & $0$ & $3 \times 10^{-5}\,\%$ \\
MLP with self-attention & no minimum & $150\,\%$ & --- \\
\bottomrule
\end{tabular}
\end{table}

On the plain MLP all three methods agree to the FD truncation level. On
the coupled models the \passthrough{\lstinline!grad\_outputs=ones!}
idiom is wrong by the full Laplacian (BN) or by more than the Laplacian
(self-attention): the silent bug of \Cref{sec:persample}. FD does not
compute the per-sample derivative either. For BN, concatenating the
stencil evaluates the network with the statistics of the \(5B\)-point
batch, so FD differentiates the function with those statistics held
fixed; the calibration accordingly selects a step an order of magnitude
larger than on the plain MLP, and the discrepancy is small at
\(B = 1024\) only because a single sample contributes \(O(1/B)\) to the
batch statistics. We do not rely on it. For self-attention, shifting
every token of a sequence by the same \(\varepsilon\) yields the
collective derivative of the whole sequence: the error of FD against the
per-sample reference has no minimum within the calibration grid, even
after the grid was extended down to \(10^{-10}\) (the argmin sits at a
grid edge and changes between initializations) and exceeds the Laplacian
itself at every step, so no calibrated FD value is reported.

For BN the coupling can be removed at the layer level. The BN variant
proposed in Algorithm 3 always uses its running statistics, so the
output of sample \(i\) depends on sample \(i\) only. In training mode
the running statistics are updated from the current batch without
gradient, as a side effect of the forward pass, and are debiased like
the moments of Adam so that the layer works from the first iteration.
With this layer batched AD is exact and FD agrees to the truncation
level at its calibrated step (third row of \Cref{tab:bn}); during a
stencil evaluation all \(5B\) points see the same statistics. We verify
the layer at the level of the derivatives only; training PINNs with it
is left for future work. We do not propose a corresponding fix for
self-attention.

\textbf{Algorithm 3: BN without inter-sample dependencies
(\passthrough{\lstinline!DecoupledBatchNorm1d!})}

\begin{lstlisting}
CLASS DecoupledBatchNorm1d(num_features, momentum m = 0.1, eps = 1e-5)
    # learnable gamma = 1, beta = 0; buffers mean = 0, var = 0, t = 0
    METHOD forward(x):                     # x: (B, num_features)
        if training:                       # running statistics, no gradient
            mean <- (1 - m) * mean + m * batch_mean(x)
            var  <- (1 - m) * var  + m * batch_var(x)
            t <- t + 1
        c <- 1 - (1 - m)^t                 # debiasing, as in Adam
        return gamma * (x - mean / c) / sqrt(var / c + eps) + beta
\end{lstlisting}

\hypertarget{app:convergence}{%
\section{Convergence curves}\label{app:convergence}}

\begin{figure}
\hypertarget{fig:convergence}{%
\centering
\includegraphics[width=1\textwidth,height=\textheight]{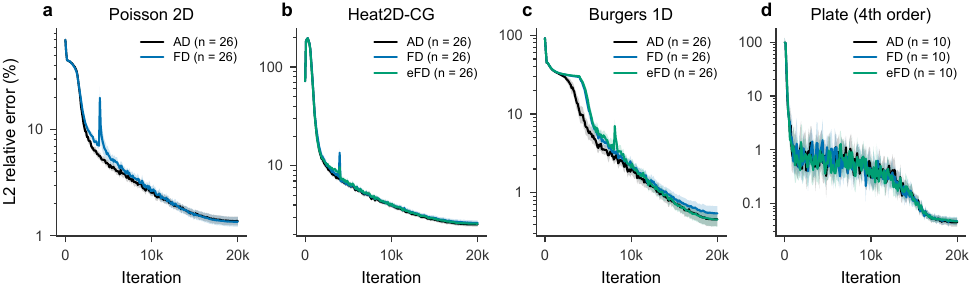}
\caption{Convergence of AD, FD and eFD on (a) Poisson 2D, (b) Heat2D-CG,
(c) Burgers 1D and (d) the plate: L2 relative error on the full
reference grid, evaluated every 100 iterations; geometric mean over
seeds and 95\% CI computed in log space. The spikes on the FD and eFD
curves at iterations 4000 and 8000 are the recalibrations of
\(\varepsilon\) (\Cref{sec:calibration}).}\label{fig:convergence}
}
\end{figure}

\citet{Chen2024Automatic} argue, from a spectral analysis of the
training dynamics, that replacing AD by finite differences slows the
optimization and raises the error floor. The tables in
\Cref{sec:experiments} report only the final error, so
\Cref{fig:convergence} shows the whole training history of the same
runs.

On all four problems FD and eFD follow AD within the confidence band for
most of the run and end at the same level. Two deviations are visible.
Recalibration of \(\varepsilon\) produces a spike that decays within a
few hundred iterations; the largest is on Poisson at iteration 4000,
where the step moves by more than a decade. On Burgers, FD and eFD lag
behind AD between iterations 4000 and 9000 and catch up by iteration
10000. We do not observe the slower optimization predicted by
\citet{Chen2024Automatic}: their analysis concerns a fixed, uncalibrated
step, and \Cref{fig:teaser}c shows where such a step would indeed harm
the result.